\documentclass[10pt,twocolumn,letterpaper]{article}

\usepackage{cvpr}              

\usepackage{graphicx}
\usepackage{amsmath}
\usepackage{amssymb}
\usepackage{booktabs}
\usepackage{multirow}
\usepackage[export]{adjustbox}
\usepackage{threeparttable}
\usepackage{subfloat}
\usepackage{tabularx}
\usepackage{adjustbox}
\usepackage{color}
\usepackage{subfig}
\usepackage[misc]{ifsym} 
\definecolor{linkcolor}{RGB}{255,0,0}
\definecolor{urlcolor}{RGB}{255,105,180}
\definecolor{citecolor}{RGB}{66,168,235}

\usepackage[pagebackref,breaklinks,colorlinks]{hyperref}
\hypersetup{colorlinks=true,linkcolor=linkcolor,urlcolor=urlcolor,citecolor=citecolor}

\usepackage[capitalize]{cleveref}
\crefname{section}{Sec.}{Secs.}
\Crefname{section}{Section}{Sections}
\Crefname{table}{Table}{Tables}
\crefname{table}{Tab.}{Tabs.}

\def\confName{CVPR}
\def\confYear{2022}

\begin{document}

\title{Video K-Net: A Simple, Strong, and Unified Baseline for Video Segmentation}

\author{
Xiangtai Li$^{1,2,4}$\thanks{Equal contribution. Most Work Done when Xiangtai was in SenseTime Research (Beijing).} \quad
Wenwei Zhang$^{2*}$ \quad
Jiangmiao Pang$^{3,5*}$ \quad
Kai Chen $^{4,5}$ \quad \\
Guangliang Cheng$^{4\textrm{\Letter}}$ \quad 
Yunhai Tong$^{1\textrm{\Letter}}$ \quad
Chen Change Loy$^{2}$
\\[0.1cm]
\small $ ^1$ Key Laboratory of Machine Perception, MOE, School of Artificial Intelligence, Peking University \\
\small $ ^2$ S-Lab, Nanyang Technological University \quad \small $ ^3$ CUHK-SenseTime Joint Lab, the Chinese University of Hong Kong \\
\small $ ^4$ SenseTime Research \quad
\small $ ^5$ Shanghai AI Lab \\
\small\texttt{\{lxtpku, yhtong\}@pku.edu.cn \quad \{wenwei001, ccloy\}@ntu.edu.sg } \\
\small\texttt{pangjiangmiao@gmail.com \quad \{chenkai, chengguangliang\}@sensetime.com}
}

\maketitle



\begin{figure*}[!t]
	\centering
	\includegraphics[width=0.97\linewidth]{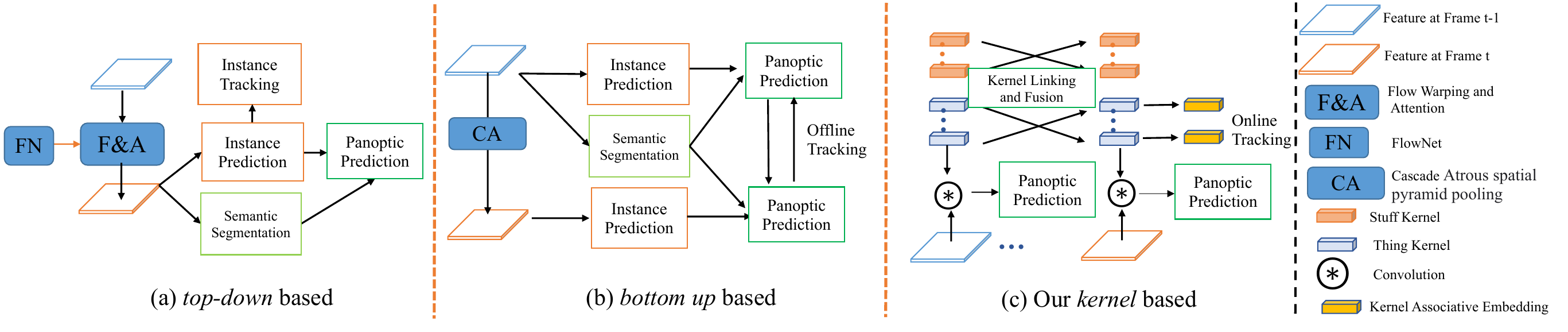}
	\caption{ \small An illustration of previous \textit{top-down} based VPS method (a), \textit{bottom-up} based VPS method (b) and the proposed Video K-Net (c). Unlike previous approaches~\cite{ViPDeepLab,kim2020vps} that perform panoptic segmentation and object tracking with independent modules, our method unifies panoptic segmentation and instance level tracking via kernels in a simpler framework. }
	\label{fig:teaser2}
\end{figure*}

\begin{abstract}
This paper presents Video K-Net, a simple, strong, and unified framework for fully end-to-end video panoptic segmentation. The method is built upon K-Net, a method that unifies image segmentation via a group of learnable kernels. We observe that these learnable kernels from K-Net, which encode object appearances and contexts, can naturally associate identical instances across video frames.
Motivated by this observation, Video K-Net learns to simultaneously segment and track ``things" and ``stuff" in a video with simple kernel-based appearance modeling and cross-temporal kernel interaction. Despite the simplicity, it achieves state-of-the-art video panoptic segmentation results on Citscapes-VPS, KITTI-STEP, and VIPSeg without bells and whistles. In particular, on KITTI-STEP, the simple method can boost almost 12\% relative improvements over previous methods. On VIPSeg, Video K-Net boosts almost 15\% relative improvements and results in 39.8 \% VPQ. We also validate its generalization on video semantic segmentation, where we boost various baselines by 2\% on the VSPW dataset. Moreover, we extend K-Net into clip-level video framework for video instance segmentation, where we obtain 40.5\% for ResNet50 backbone and 54.1\% mAP for Swin-base on YouTube-2019 validation set. We hope this simple, yet effective method can serve as a new, flexible baseline in unified video segmentation design. Both code and models are released at \url{https://github.com/lxtGH/Video-K-Net.}
\end{abstract}


\section{Introduction}
\label{sec:intro}

Video Panoptic Segmentation (VPS) aims at segmenting and tracking every pixel of input video clips~\cite{kim2020vps,STEP,hurtado2020mopt}. As a fundamental technique to scene understanding, it has received increasing attention in recent years due to its wide applications in many vision systems, including autonomous driving and robot navigation~\cite{geiger2012we,cordts2016cityscapes}.
By definition, VPS is an extension of Panoptic Segmentation (PS)~\cite{kirillov2019panoptic} into the video domain with the goal of unifying Video Semantic Segmentation (VSS)~\cite{shelhamer2016clockwork,DFF,miao2021vspw} and Video Instance Segmentation (VIS)~\cite{vis_dataset,mask_pro_vis} into a single task.

\begin{figure}[!t]
	\centering
	\includegraphics[width=1.0\linewidth]{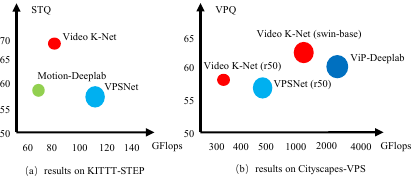}
	\caption{ \small Comparison with previous work. GFLOPS are obtained by using the original image inputs of KITTI-STEP ($384 \times 1280$) and Cityscapes-VPS ($1024 \times 2048$). The number of network parameters are indicated by the radius of circles. Note that our Video K-Net obtains higher accuracy while having lower model complexity on Cityscapes-VPS. Compared to Motion-Deeplab~\cite{STEP}, our method achieves a significant gain (10\%) with a few extra GFLOPS. }
	\label{fig:teaser1}
\end{figure}

Existing studies for video panoptic segmentation can be mainly divided into \textit{top-down} methods~\cite{kim2020vps,hurtado2020mopt} and \textit{bottom-up} approaches~\cite{ViPDeepLab,STEP}.
Top-down methods try to solve VPS as a multi-task learning problem via performing semantic segmentation, instance segmentation and multiple object tracking individually. VPSNet~\cite{kim2020vps} is proposed to learn to fuse and warp features. In particular, it applies an optical flow network~\cite{FlowNet} to align features and attention modules to fuse features~\cite{woo2018cbam}. However, these approaches~\cite{kim2020vps,hurtado2020mopt} involve many hyper-parameters and ad-hoc designs, leading to a complex system. Bottom-up methods~\cite{cheng2020panoptic,ViPDeepLab,STEP} first perform semantic segmentation and then predict near frames' centers to localize and track each instance where both features are connected by an ASPP module~\cite{deeplabv3} (Fig.~\ref{fig:teaser2}-(b)). Despite being simpler than top-down approaches, they rely on several post-processing components (i.e., NMS for center grouping, offline mask tracking).

Recently, inspired by the design of object queries in Detection Transformer (DETR)~\cite{detr}, new efforts~\cite{wang2020maxDeeplab,zhang2021knet,cheng2021maskformer} emerge to explore unified solutions to image panoptic segmentation.
In particular, MaskFormer~\cite{cheng2021maskformer} and K-Net~\cite{zhang2021knet} unify `things' and `stuff' segmentations by dynamic kernels within the mask classification paradigm. 
The aforementioned studies in image segmentation motivate us to simplify cumbersome pipelines in video panoptic segmentation. 
It is noteworthy that several studies~\cite{meinhardt2021trackformer,zeng2021motr,transtrack} in Multi Object Tracking (MOT) also adopt the query design to achieve end-to-end learning. However, most of them introduce \textit{extra tracking queries or boxes} to handle each tracked instance. 


In this paper, we present Video K-Net, a fully end-to-end framework for video panoptic segmentation.
It is observed that the learnable kernels from K-Net~\cite{zhang2021knet}, which encode object appearances and contexts, are naturally capable of directly associating identical objects across videos. 
These kernels can be directly used to extract discriminative embeddings for object tracking, which can potentially simplify the MOT framework.

To this end, we design a simple embedding head with a modified contrastive learning loss to obtain better temporal association embeddings. We directly take the dynamic kernels as inputs and output the \textit{kernel association embeddings}. Such design avoids extra tracking query design. The embedding loss forces the kernels to be more distinctive. Then we also link the kernels between adjacent frames for better and more consistent tracking results. Moreover, to jointly learn more consistent temporal mask classification and prediction, we also propose a temporal kernel fusion module to fuse the kernels with corresponding kernel features in a more efficient manner. Fusing at kernel level avoids computation cost at feature levels (Attention in Fig.~\ref{fig:teaser2}-(a) and Cascaded ASPP in Fig.~\ref{fig:teaser2}-(b)). In this way, as shown in Fig.~\ref{fig:teaser2}-(c), the VPS task can be seen as a kernel linking problem and can significantly reduce the pipeline complexity without extra tracking query and RoI features. The tracking is performed in an online manner. Moreover, it only adds a few extra GFLOPs compared to original K-Net (2.3\%GFLOPs). Different from previous works that use RoI heads~\cite{kim2020vps,qdtrack}, our kernel based embedding learning avoids box cropping and results in better association performance. 

Video K-Net obtains consistent and significant improvements over the strong baseline K-Net with \textbf{3.5\% STQ} on KITTI-STEP and \textbf{4\% VPQ} on Cityscapes-VPS. In particular, our method boosts almost \textbf{12\%} relative improvements over previous bottom up baseline Motion-Deeplab~\cite{STEP}. On VIPSeg dataset~\cite{miao2022large}, our Video K-Net outperforms the co-current baseline CLIP-PanoFCN~\cite{miao2022large} by \textbf{3\% VPQ} and boosts almost \textbf{16\%} relative improvements over CLIP-PanoFCN.
Video K-Net achieves new state-of-the-art results on three VPS datasets, including KITTI-STEP~\cite{STEP}, Cityscapes-VPS~\cite{kim2020vps}, and VIPSeg~\cite{miao2022large}. Moreover, as shown in Fig.~\ref{fig:teaser1}, Video K-Net achieves the best trade-off between accuracy and GFLOPs on the two datasets. We further validate the effectiveness of kernel fusion on VSS task with VSPW dataset~\cite{miao2021vspw} and we boost previous baselines~\cite{deeplabv3plus,zhao2017pyramid} via considerable margins on two different metrics.  

Moreover, we also extend the Video K-Net into clip-level processing via extra temporal kernel fusion module for VIS task~\cite{vis_dataset}. We achieve 40.5\% AP using ResNet50 backbone, which lead to \textbf{4.0\%} improvements over previous VisTR~\cite{VIS_TR} with less GFLOPs. In summary, extensive experiments and analysis on three different types of video segmentation tasks demonstrate that Video K-Net can serve as a new baseline for future research on \textit{unified video segmentation tasks}. 
\section{Related Work}
\label{sec:relatedwork}

\noindent
\textbf{Panoptic Segmentation.} 
The goal of this task is to unify the semantic segmentation and instance segmentation into one framework with a single metric named Panoptic Quality (PQ)~\cite{kirillov2019panoptic}. Since then, lots of works~\cite{xiong2019upsnet,kirillov2019panopticfpn,li2020panopticFCN,chen2020banet,li2018learning,porzi2019seamless,yang2019sognet,Wu2020AutoPanopticCM,cheng2020panoptic,axialDeeplab} have been proposed to solve this task with various approaches. However, most methods separate thing and stuff segmentation as individual tasks. Recently, starting from DETR~\cite{detr}, query based approaches~\cite{wang2020maxDeeplab,cheng2021maskformer,zhang2021knet,panopticpartformer} unify both things and stuff segmentation as a set prediction problem. In particular, K-Net~\cite{zhang2021knet} unifies segmentation tasks in the view of dot product between kernel and feature maps.

\noindent
\textbf{Video Semantic/Instance Segmentation.} Video Semantic Segmentation, a direct extension of semantic segmentation to the video scenario, requires predicting a semantic label to every pixel in each video frame. Several approaches~\cite{shelhamer2016clockwork,DFF,hu2020tdnet,li2021improving} have been proposed in the literature mainly to model the temporal association such as optical flow warping or attention. Recently, the VSPW dataset~\cite{miao2021vspw} is proposed to evaluate large scale video semantic segmentation in the wild. Video Instance Segmentation (VIS)~\cite{vis_dataset} extends instance segmentation into video, and it aims to simultaneously classify, segment and track object instances in a given video sequence. Several methods~\cite{mask_pro_vis,lin2021video,zhou2022transvod} are proposed to link instance-wised feature in video. Recently, several works~\cite{VIS_TR,hwang2021video} extend DETR into VIS. Multi-Object Tracking and Segmentation (MOTS) task~\cite{voigtlaender2019mots} is proposed to evaluate Multi-Object Tracking along with instance segmentation. However, both VIS and MOTS have limited scale distribution and much fewer objects in the scene. 

\noindent
\textbf{Video Panoptic Segmentation.} Video Panoptic Segmentation (VPS)~\cite{kim2020vps,hurtado2020mopt} requires generating the instance tracking IDs along with panoptic segmentation results across video clips. Kim~\etal~\cite{kim2020vps} use Cityscapes video sequences for 6 frames out of each short 30 frame clip and mainly focus on short-term tracks. They proposed Video Panoptic Quality (VPQ) for evaluation. Several works~\cite{kim2020vps,woo2021learning_associate_vps} are proposed to solve this task respectively.  VIP-Deeplab~\cite{ViPDeepLab} extends the Panoptic-Deeplab~\cite{cheng2020panoptic} with next frame center map prediction for DVPS task~\cite{yuan2021polyphonicformer}. Then they use such predicted maps for offline tracking. Since Cityscapes VPS only contains short term clips, to allow long term VPS, STEP dataset~\cite{STEP} is proposed. They propose a new metric named Segmentation and Tracking Quality (STQ) that decouples the segmentation and tracking error. They also provide several baselines for reference. Our Video K-Net is verified to work well on both short- and long-term videos.

\noindent
\textbf{Object Tracking.} One of the major tasks in VPS is object tracking. Many studies adopt the tracking-by-detection paradigm~\cite{bewley2016simple,leal2016learning,schulter2017deep,sharma2018beyond,xu2019spatial,zhu2018online,porzi2020learning} and they divide the task into two subtasks where an object detector finds all objects and then a tracking algorithm is employed to associate them. There are also several works~\cite{zhang2018integrated,bergmann2019tracking,peng2020chained,JDE,qdtrack,zhou2020tracking} that detect and track objects at the same time. There are also tracking methods using object queries. For VPS, ViP-DeepLab~\cite{ViPDeepLab} performs object tracking by clustering all instance pixels. By contrast, our method directly extends and links kernels into kernel association embeddings and avoids such complex post-grouping steps. 

\begin{figure}[h!]
	\centering
	\includegraphics[width=1.0\linewidth]{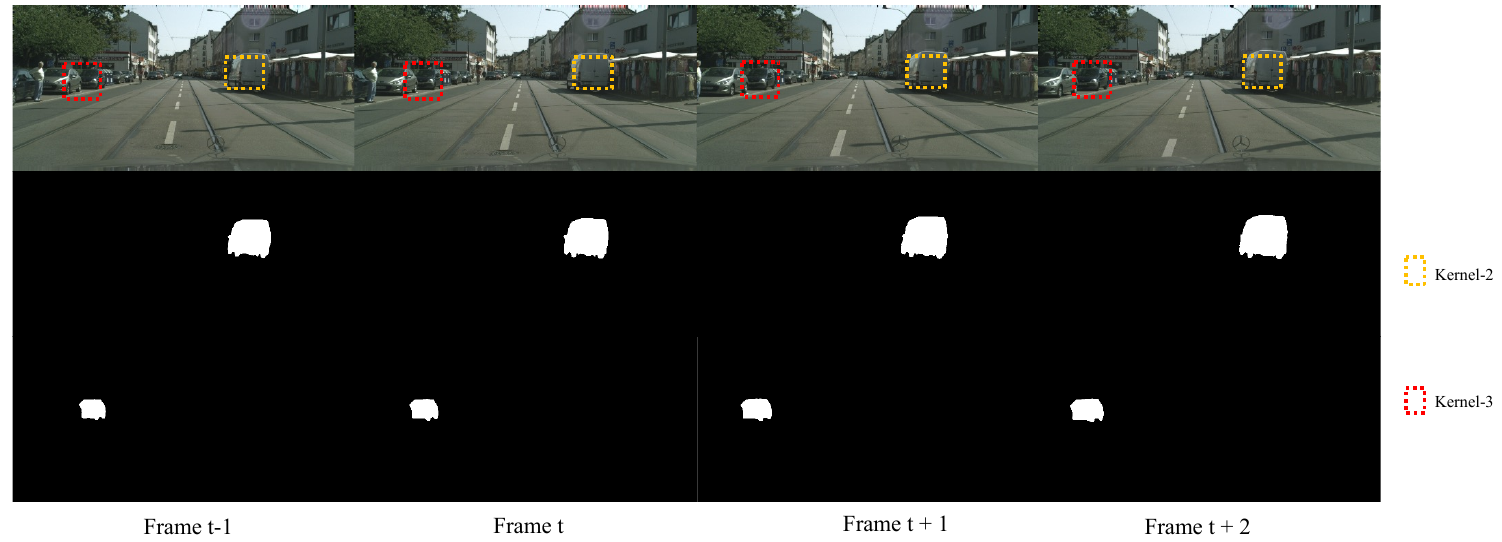}
	\caption{\small Toy experiment illustration. We use the K-Net directly on Cityscapes video datasets. We find that several instances are originated from \textbf{the same kernel} predictions (Red, Yellow boxes, \textcolor{yellow}{Kernel-2} and \textcolor{red}{Kernel-3}). This observation motivates us to use K-Net directly on video. Best view it in color.}
	\label{fig:toy_exp}
\end{figure}

\section{Method}
\label{sec:method}


In this section, we will overview the image baseline K-Net. Then we present several toy experiments on using K-Net in video without extra tracking components. Motivated by that, we propose a simple yet effective approach to learn the kernel association embeddings from kernels and introduce two improvements via linking and fusing kernels on K-Net. Finally, we detail the entire framework including both training and inference.

\subsection{Using K-Net on Video}

\begin{table}[t]
  \centering
    \caption{\textbf Toy Experiment results on KITTI-STEP and Cityscape-VPS set with $STQ$ and $VPQ$ metrics. Unitrack~\cite{wangUnitrack} uses ResNet-50 as the appearance model. }
  \scalebox{0.70}{
  \begin{tabular}{l c c c c c }
    \toprule[0.2em] 
    \textbf{KITTI-STEP} & Backbone & STQ & AQ & SQ & - \\
    \toprule[0.2em]
    K-Net & ResNet50 &  67.5 & 65.5 & 68.9 & -  \\
    K-Net + Unitrack~\cite{wangUnitrack} & ResNet50  & 65.1 & 64.3 & 68.9 & - \\
    \toprule[0.2em]
    \textbf{Cityscapes-VPS} & Backbone & - & - & - & VPQ \\
    \toprule[0.2em]
    K-Net & ResNet50 & - & - & -& 54.3  \\
    K-Net + Unitrack~\cite{wangUnitrack} & ResNet50  & - & - & - & 53.2 \\
    \bottomrule[0.2em]
  \end{tabular}
  }
  \label{tab:toy_exp}
\end{table}

\begin{figure*}[t!]
	\centering
	\includegraphics[width=0.90\linewidth]{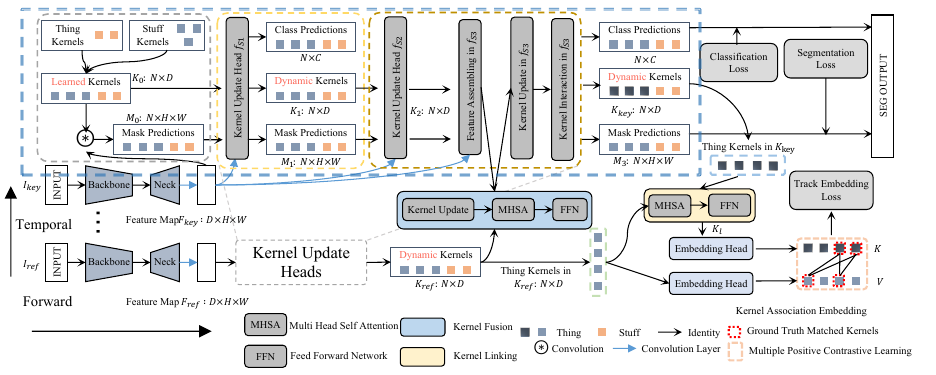}
	\caption{\small An illustration of our proposed Video K-Net. Our method is based on K-Net~\cite{zhang2021knet}(in blue dashed box), which is the top-left part of the figure. Video K-Net adds Kernel Fusion at the start phase of the last stage. The Kernel Linking is performed on the output of dynamic kernels. The Embedding Head is appended at the output of kernel linking and takes kernel outputs from both sampled frames.}
	\label{fig:methods}
\end{figure*}

\noindent 
\textbf{Overview of K-Net.} K-Net~\cite{zhang2021knet} formulates different image segmentation tasks (semantic, instance, panoptic) into a unified framework via a group of learnable kernels, where each kernel is responsible for generating a mask for either a potential instance or a stuff class. Started with a set of randomly initialized convolutional kernels, the goal of K-Net is to learn kernels in accordance to the segmentation targets (thing or stuff masks). To distinguish and separate various instances, a cascaded decoder is proposed to refine the learned kernels iteratively. In particular, the authors propose a kernel update strategy that enables each kernel to be dynamic and conditional on its meaningful area of the input features. As shown in the top-left of Fig.~\ref{fig:methods}, it mainly contains three steps: 1, group feature assembling to obtain corresponding kernel features (in grey color box). 2, adaptive feature update, where the assembled kernel features are used to update the learned kernel (in yellow color box). 3, kernel interaction, where the learned kernels exchange contextual information via self attention layers~\cite{vaswani2017attention} (in brown color box). Then the final dynamic kernels are used to perform mask classification via fully connected layers. The mask segmentation is done via inner product between the updated kernels and input position-aware features. 
Then these predictions can be trained in an end-to-end manner with bipartite matching. The above process is repeated several times and the last stage outputs are chosen as the final outputs.

\noindent
\textbf{Toy Experiments on using K-Net on Video.} 
Before detailing the proposed Video K-Net, we present several toy experiments to show the motivation of our work. As mentioned in the previous section, refining kernels is the key for making K-Net work. We first depict kernel indexes of a trained K-Net on Cityscapes video sequence~\cite{cordts2016cityscapes} in Fig.~\ref{fig:toy_exp}. We use K-Net directly into video segmentation task. We argue that the \textit{updated kernels contain discriminative information}, which can be directly used to track the instance in video even \textit{without} adding extra tracking heads. The insights are as follows: 1, Each output instance mask corresponds with one specific kernel. 2, Each kernel absorbs the position-aware features during the adaptive feature update, where the instance aware information has already merged into each kernel. Thus, the same instance can be decoded from the previous kernel. In Fig.~\ref{fig:toy_exp}, we present one visual examples on Cityscapes-VSP. We train the K-Net directly on both datasets \textit{without} adding tracking components. As shown in that figure, in three frames, several kernels are well linked, such as person and car. 

Moreover, we take a further step by directly using the learned kernels as appearance embedding for tracking association. In particular, we use the Quasi-Dense Tracker~\cite{qdtrack} which is a pure appearance based tracker. We replace the appearance embeddings with our learned kernels from the last stage. As shown in Tab.\ref{tab:toy_exp}, surprisingly, we found the performance of original K-Net is already good enough to perform tracking and even achieves better results than Unitrack~\cite{wangUnitrack} that comes with extra appearance network and motion prediction module~\cite{deepsort}. These findings motivate us to dive into the design of linking kernels along the temporal dimension.  

\subsection{Extending K-Net into Video}

Despite the competitive performance of the original K-Net, it shows several failure cases, for instance, the fast motion object shown in Fig.~\ref{fig:toy_exp}. Thus, we design three improvements on K-Net including the learning of kernel association embeddings via a modified contrastive learning loss, learning to link tracking kernels, and learning to fuse kernels, respectively.

As shown in Fig.~\ref{fig:methods}, given a key image $I_{key}$ for training, we randomly select a reference image $I_{ref}$ from its temporal neighborhood. The neighbor distance is constrained by a window size ranging from $ [-2, 2]$ in our experiments by default. Then we use K-Net to obtain the two individual kernels: $K_{key}$ and $K_{ref}$.

\noindent
\textbf{Learning Kernel Association Embeddings.} Our method is motivated by the recent work Quasi-Dense~\cite{qdtrack} in MOT. Instead of using RoI boxes for feature embedding extraction, we use the kernels directly. We add an extra lightweight embedding head after the original K-Net decoder, to extract embedding features for each kernel. The embedding head is implemented via several fully connected layers. We adopt mask-based assignment for Quasi-Dense learning. A kernel embedding is defined as positive to an object if their corresponding masks have an IoU higher than $\alpha_1$, or negative if their corresponding masks have an IoU lower than $\alpha_2$. The values of $\alpha_1$ and $\alpha_2$ are set as 0.7 and 0.3 in our experiments. The matching of kernels on two sampled frames is positive if the two regions are associated with the same object, and negative otherwise. Instead of following the original design~\cite{qdtrack} that uses hundreds of proposals during training, we \textbf{\textit{only optimize kernels that are matched}} with ground truth masks. This is because most kernels (unmatched) are not accurate and may cause noise during training, leading to inferior results. 

Assume there are $V$ matched kernels on the key frame as training samples and $K$ matched kernels on the reference frame as contrastive targets, where both $V$ and $K$ are always \textit{fewer than} kernel number $N$. As shown in red boxes of the Fig.~\ref{fig:methods}, these kernels are sampled from $K_{key}$ and $K_{ref}$, respectively. 
\begin{align}
    \label{equ:qd_loss}
    \mathcal{L}_\text{track} & = -\sum_{\textbf{k}^{+}}\text{log}
    \frac{\text{exp}(\textbf{v} \cdot \textbf{k}^{+})}
    {\text{exp}(\textbf{v} \cdot \textbf{k}^{+}) + \sum_{\textbf{k}^{-}}\text{exp}(\textbf{v} \cdot \textbf{k}^{-})},
\end{align}
where $\textbf{v}$, $\textbf{k}^{+}$, $\textbf{k}^{-}$ are kernel embeddings of the training sample, its positive targets, and negative targets in $K$. In addition, following previous work~\cite{qdtrack}, we also adopt L2 loss as an auxiliary loss.

\begin{equation}
    \label{equ:qd_loss_aux}
    \mathcal{L}_\text{aux} = (\frac{\textbf{v} \cdot \textbf{k}}{||\textbf{v}|| \cdot ||\textbf{k}||} - c)^2,
\end{equation}
where $c$ is 1 if the match of two samples is positive and 0 otherwise. In comparison to the original work, our modification makes the loss calculation process sparse, benefiting the training process and achieving better results. (See the experiment part in Sec.~\ref{sec:ablation}). In summary, we term this procedure as learning Kernel Association Embedding (KAE).

\noindent 
\textbf{Learning to Link Kernels.} Beyond supervision, we take a further step to link kernels $K_{key}$ and $K_{ref}$ during training and inference. This forces the kernels to perform interaction along the temporal dimension. We adopt one self-attention layer ($MHSA$: Multi Head Self Attention) with a Feed Forward Network ($FFN$)~\cite{vaswani2017attention} to learn the correspondence among each query to obtain the updated queries, allowing the full correlation among queries. 
This process is shown as follows: 
\begin{equation}
    K_{l} = FFN(MHSA({K}_{key},{K}_{ref},{K}_{ref}) 
    + {K}_{key}),
 \label{equ:selfattention}
\end{equation}
where the query, key and value are ${K}_{key}$, ${K}_{ref}$ and ${K}_{ref}$, respectively. In this way, kernels from the reference frame are propagated into key frame via the affinity matrix between kernels. Then the linked kernel $K_{l}$ is sent to the tracking head. During the training, this operation is \textit{only optimized with matched thing kernels}. During the inference stage, this operation is performed on \textit{final preserved kernels} in panoptic segmentation map, where other thing kernels are dropped. 

\noindent
\textbf{Learning to Fuse Kernels.} The previous linking step may focus on the tracking consistency while ignoring the segmentation consistency. To address this issue, we propose to fuse kernels in between frames on K-Net. In particular, as shown in Fig.~\ref{fig:methods},  we use input kernels of the last stage from two frames. Then we perform kernel updating, where we use kernel features of the current frame to update the previous kernel (Step-2 from original K-Net). Then we adopt one self-attention layer with feed forward layers~\cite{vaswani2017attention} to fuse the previous kernels into the current frame. The updating step is essential. Directly fusing kernels leads to bad results, since large motion and scale variation often occur in the video scene. More details can be found in Sec.~\ref{sec:ablation}. In summary, through the above steps, we link kernels of K-Net into the video domain with a little extra computation cost. 

\subsection{Video K-Net Architecture}
\noindent
\textbf{Network Architecture.} By default, we adopt the panoptic segmentation setting of K-Net. The kernels are composed of instance kernels and semantic kernels for thing and stuff mask prediction. We adopt semantic FPN~\cite{kirillov2019panopticfpn} for producing high resolution feature map with extra positional encoding as in~\cite{vaswani2017attention,detr}. The kernel linking operation is appended at the last stage of the K-Net decoder. The kernel fusion is placed at the beginning part. The lightweight tracking head is appended after the decoder. Note that backbone, neck, update head, and embedding head are shared across frames. For VSS task, we directly append the head of Video K-Net at the end of existing semantic segmentation network~\cite{deeplabv3plus}. 

\noindent
\textbf{Loss Function and Training.} Compared to the original K-Net, we add the tracking loss at the end of K-Net loss. The total loss function for all kernels is shown as
$L = \lambda_{cls}L_{cls} + \lambda_{ce}L_{ce} + \lambda_{dice}L_{dice} + \lambda_{track}L_{track} + \lambda_{aux}L_{aux}$ ,
where $L_{cls}$ is Focal loss~\cite{focal_loss} for classification,
and $L_{ce}$ and $L_{dice}$ are Cross Entropy (CE) loss and Dice loss~\cite{dice_loss,wang2019solo} for segmentation, respectively. $L_{track}$ and $L_{aux}$ are the tracking loss for instance kernels only.  We adopt Hungarian assignment strategy used in~\cite{detr} for target assignment for end-to-end training. It builds a one-to-one mapping between the predicted instance masks and the ground-truth instances based on the masked-based matching costs. During training, we found that applying the segmentation loss on both key frame and reference frames leads to better results.

\noindent
\textbf{Inference.} We use K-Net to generate panoptic segmentation results, where we paste thing and stuff masks in a mixed order. We use the kernel embeddings from the learned embedding head as association features. We take these features as the input of Quasi-Dense Tracker~\cite{qdtrack} where the bidirectional softmax is calculated to associate the instances between two frames in an online manner. Noted that we only track the preserved instance masks from panoptic segmentation maps to save computation cost. We also compare our tracking algorithm with the original Quasi-Dense Tracker~\cite{qdtrack} using the boxes for appearance modeling.

\noindent
\textbf{Clip-level Training and Inference for VIS.} We also extend our Video K-Net into Clip-level training and inference pipeline for Video Instance Segmentation~\cite{vis_dataset}. In particular, we add three kernel fusion layers on the top of K-Net outputs to jointly fuse both kernels and the corresponding kernel features along the temporal dimension. We use the mean of temporal kernels to represent each object for each clip and assign instance ID to each instance in each frame via kernel index~\cite{VIS_TR}.  

\section{Experiment}
\label{sec:experiment}
%

\subsection{Experiment Setup}

\noindent
\textbf{Dataset.} We carry out experiments on four video-level datasets: KITTI-STEP, Cityscaeps-VPS, VSPW, VIPSeg, and YouTube-VIS. KITTI-STEP has 21 and 29 sequences for training and testing, respectively. The training sequences are further split into a training set (12 sequences) and a validation set (9 sequences). Cityscapes-VPS contains 400 training, 50 validation, and 50 test videos. Following previous work~\cite{kim2020vps}, all 30 frames are predicted during the inference, and only 6 frames with ground truth are evaluated. Both Cityscapes-VPS and KITTI-STEP have the same class number with Cityscapes dataset~\cite{cordts2016cityscapes}. However, the definition of thing and class is different. For VSPW dataset and YouTube-VIS dataset, refer to \textit{the appendix file}.


\noindent
\textbf{Implementation Details.} We implement our models in PyTorch~\cite{pytorch_paper} with MMDetection toolbox~\cite{chen2019mmdetection}. We use the distributed training framework with 8 GPUs. Each mini-batch has one image per GPU. Following previous work, we first pretrain the image baseline on Cityscapes dataset~\cite{cordts2016cityscapes} where we adopt the same pretraining setting~\cite{cheng2020panoptic,STEP} for fair comparison. For STEP pretraining, we change the class distribution of Cityscapes dataset into STEP format to avoid overfitting on STEP. Both ResNet~\cite{resnet} and Swin Transformer~\cite{liu2021swin} are adopted as the backbone networks, and other layers use Xavier initialization~\cite{xavier_init}. The optimizer is AdamW~\cite{ADAMW} with weight decay 0.0001. We adopt full image size for random crop in both pretraining and training process. For the large Swin Transformer backbone~\cite{liu2021swin}, we also pretrain our model on Mapillary dataset~\cite{neuhold2017mapillary}. More details of pretraining and finetuning can be found in the appendix file. Due to diversity of dataset, for VSPW, YouTube-VIS and VIP-Seg dataset, refer to \textit{the appendix file}.


\begin{table}[t]
  \centering
   \caption{\small \textbf{Experiment results on KITTI set with both $STQ$ and $VPQ$ metric.} OF refers to an optical flow network~\cite{teed2020raft}. The results on validation set are shown in the several top rows, and results on test set are in the bottom rows. P means Panoptic Deeplab~\cite{cheng2020panoptic}. Following~\cite{STEP}, we keep two decimal numbers. $VPQ$ is obtained via average results of window size $k$ where $k=1,2,3,4$~\cite{STEP}. Top: validation set. Bottom: test set. We find 0.5\% noise on this dataset, where we report the average results (three times). Axial means using extra Axial Attention~\cite{axialDeeplab}.}
  \scalebox{0.60}{
  \begin{tabular}{l c c || c c c c }
    \toprule[0.2em]
    \textbf{KITTI-STEP} & Backbone & OF & STQ & AQ & SQ & VPQ \\
    \toprule[0.2em]
    P + IoU Assoc. & ResNet50 & & 0.58 & 0.47 & 0.71 & 0.44  \\
    P + SORT & ResNet50 &  & 0.59  & 0.50 & 0.71 & 0.42  \\
    P + Mask Propagation & ResNet50 & \checkmark & 0.67 & 0.63 & 0.71 & 0.44 \\
   
    Motion-Deeplab~\cite{STEP}& ResNet50 &  & 0.58 & 0.51 & 0.67 & 0.40  \\
    VPSNet~\cite{kim2020vps}& ResNet50  & \checkmark & 0.56 & 0.52 & 0.61 & {0.43}  \\
    TubeFormer-DeepLab~\cite{kim2022tubeformer} & ResNet-50 + axial &  & 0.70 & 0.64 &  0.76 & 0.51 \\
    \hline
    Video K-Net & ResNet50 &  & 0.71 & 0.70  & 0.71  &  0.46 \\
    Video K-Net & Swin-base &  & 0.73 & 0.72 & 0.73 & 0.53 \\
    Video K-Net & Swin-large & & 0.74 & 0.73 & 0.75 & 0.56 \\
    \bottomrule[0.2em]
    Motion-Deeplab~\cite{STEP}& ResNet50 &  & 0.52 & 0.46 & 0.60 & -  \\
    \hline
    Video K-Net & ResNet50 &  & 0.59 & 0.50  & 0.62  &  - \\
    Video K-Net & Swin-base &  & 0.63 & 0.60 & 0.65 & - \\
    \bottomrule[0.2em]
  \end{tabular}
  }
 
  \label{tab:kitti_baselines}
\end{table}
 
\noindent
\textbf{Evaluation Metrics.} For VPS task, as discussed in STEP~\cite{STEP}, different metrics result in a significant gap even on the same VPS dataset since different metrics emphasize different properties. We adopt two widely used metrics: Video Panoptic Quality ($VPQ$) and Segmentation and Tracking Quality ($STQ$). The former mainly focuses on mask proposal level as PQ~\cite{kirillov2019panoptic} with different window sizes and threshold parameters, while the latter emphasizes pixel level segmentation and tracking without any thresholds. $STQ$ contains geometric mean of two items: Segmentation Quality ($SQ$) and Association Quality ($AQ$) where $ STQ = (SQ \times AQ)^{\frac{1}{2}} $. The former evaluates the pixel-level tracking while the latter evaluates the pixel-level segmentation results in a video clip. Due to the interpretability~\cite{STEP} of $STQ$, we adopt it for ablation studies. We also report $VPQ$ for the benchmark comparison on both datasets. For VSS task, we report the Mean Intersection over Union (mIoU) and mean Video Consistency ($mVC$)~\cite{miao2021vspw} for reference.

\begin{table*}[t]
  \centering
   \caption{\small \textbf{Results on Cityscapes-VPS validation set}. $k$ is temporal window size in~\cite{kim2020vps}. All the methods use the single scale inference without other augmentations in the test stage. In each cell, we report $VPQ$, $VPQ_{thing}$ and $VPQ_{stuff}$ in order. There is about 0.5\% noise on this dataset, where we report the average results (three times).}
  \scalebox{0.65}{
  \begin{tabular}{l|c|c|c|c|c|c}
    \toprule[0.2em]
    Method & Backbone &k = 0 & k = 5 & k = 10 & k = 15 & Average\\
    \toprule[0.2em]
    VPSNet~\cite{kim2020vps} & ResNet50 & 65.0 $\vert$ 59.0 $\vert$ 69.4 &  57.6$\vert$ 45.1 $\vert$ 66.7 & 54.4$\vert$ 39.2$\vert$ 65.6 & 52.8 $\vert$ 35.8 $\vert$ 65.3 & 57.5 $\vert$ 44.8 $\vert$ 66.7 \\
    SiamTrack~\cite{woo2021learning_associate_vps} & ResNet50 & 64.6 $\vert$ 58.3 $\vert$ 69.1 & 
    57.6 $\vert$ 45.6 $\vert$ 66.6 & 54.2 $\vert$ 39.2 $\vert$ 65.2 & 52.7 $\vert$ 36.7 $\vert$ 64.6 & 57.3 $\vert$ 44.7 $\vert$ 55.0
    \\
    ViP-Deeplab~\cite{ViPDeepLab} & WideResNet41~\cite{zagoruyko2016wideresnet} & 68.2 $\vert$ N/A $\vert$ N/A  & 61.3  $\vert$  N/A $\vert$ N/A  &  58.2 $\vert$ N/A $\vert$ N/A & 56.2  $\vert$  N/A $\vert$ N/A  &   60.9 $\vert$  N/A $\vert$ N/A \\
    ViP-Deeplab~\cite{ViPDeepLab} & WideResNet41~\cite{zagoruyko2016wideresnet}+RFP~\cite{qiao2021detectors} + AutoAug~\cite{cubuk2018autoaugment} &  69.2 $\vert$ N/A $\vert$ N/A   & 62.3  $\vert$ N/A  $\vert$ N/A &  59.2 $\vert$ N/A $\vert$ N/A  & 57.0  $\vert$  N/A $\vert$ N/A &  61.9 $\vert$ N/A  $\vert$ N/A  \\
    \midrule
    
    Video K-Net & ResNet50  & 65.6 $\vert$ 57.4 $\vert$ 71.5 & 57.7 $\vert$ 43.4 $\vert$ 68.2 & 54.2 $\vert$ 36.5 $\vert$ 67.1 & 52.3 $\vert$ 33.1 $\vert$ 66.3 & 57.8 $\vert$ 45.0 $\vert$ 66.9 \\
    Video K-Net & Swin-base~\cite{liu2021swin} & 69.2 $\vert$ 63.6 $\vert$ 73.3 & 62.0 $\vert$ 51.1 $\vert$ 70.0 & 58.4 $\vert$ 44.7 $\vert$ 68.3 & 55.8 $\vert$ 39.8 $\vert$ 67.5 & 61.2 $\vert$ 49.6 $\vert$ 69.5 \\
    
    Video K-Net & Swin-base + RFP~\cite{qiao2021detectors} & 70.8 $\vert$ 63.2 $\vert$ 76.3 & 63.1 $\vert$ 49.3 $\vert$ 73.2 & 59.5 $\vert$ 43.4 $\vert$ 72.0 & 56.8 $\vert$ 37.0 $\vert$ 71.1 & 62.2 $\vert$ 49.8 $\vert$ 71.8 \\
    \bottomrule[0.1em]
    \end{tabular}
  }
  \label{tab:city_vps_results}
\end{table*}

\begin{table}[t]
  \centering
    \caption{\small \textbf{Results on VSPW validation set}. $mVC_{c}$ means that a clip with $c$ frames is used. All methods use the same setting for fair comparison.}
  \scalebox{0.70}{
  \begin{tabular}{l c c c c c }
    \toprule[0.2em]
    \textbf{VPSW} & Backbone & mIoU & $mVC_{8}$ &$mVC_{16}$  \\
    \toprule[0.2em]
    DeepLabv3+~\cite{deeplabv3plus} & ResNet101 & 35.7 & 83.5 & 78.4 \\
    PSPNet+~\cite{zhao2017pyramid} & ResNet101 & 36.5 & 84.4 & 79.8 \\
    TCB(PSPNet)~\cite{miao2021vspw} & ResNet101 & 37.5 & 86.9 & 82.1  \\
    \hline
    Video K-Net (Deeplabv3+) & ResNet101  & 37.9 & 87.0 & 82.1 \\
    Video K-Net (PSPNet) & ResNet101  & 38.0 & 87.2  & 82.3 \\
    \bottomrule[0.2em]
  \end{tabular}
  }

  \label{tab:vspw}
\end{table}

\begin{table}[!t]
	\centering
	\caption{\small \textbf{Results on Video instance segmentation} AP (\%) on the YouTube-VIS-2019~\cite{vis_dataset} validation dataset. * means using deformable fpn~\cite{zhu2020deformabledetr}. Axial means using extra Axial Attention~\cite{wang2020maxDeeplab}.
	The compared methods are listed by publication date. }
	\label{tab:sota2019_vis}
  \scalebox{0.65}{
\begin{tabular}{ r|c|c|cccc}
\toprule[0.15em]
 Method&backbone& AP & $\rm AP_{50}$ & $\rm AP_{75}$ & $\rm AR_{1}$ & $\rm AR_{10}$  \\
\toprule[0.15em]
FEELVOS~\cite{voigtlaender2019feelvos}&ResNet50&26.9&42.0&29.7&29.9&33.4  \\
MaskTrack R-CNN~\cite{vis_dataset}&ResNet50&30.3&51.1&32.6&31.0&35.5  \\
{MaskProp\cite{mask_pro_vis}}&ResNet-50&40.0&-&42.9&-&-  \\
{MaskProp~\cite{mask_pro_vis}}&ResNet101 &42.5&-&45.6&-&-  \\
{STEm-Seg~\cite{Athar_Mahadevan20ECCV}}&ResNet50&30.6&50.7&33.5&31.6&37.1  \\
{STEm-Seg~\cite{Athar_Mahadevan20ECCV}}&ResNet101&34.6&55.8&37.9&34.4&41.6  \\
CompFeat~\cite{fu2021compfeat} & ResNet50  & 35.3 & 56.0 & 38.6 & 33.1 & 40.3  \\
{VisTR~\cite{VIS_TR}}&ResNet50& 36.2& 59.8& 36.9& 37.2 & 42.4  \\
{VisTR~\cite{VIS_TR}} &ResNet101&40.1& 64.0& 45.0 & 38.3 & 44.9  \\
TubeFormer-DeepLab~\cite{kim2022tubeformer} & ResNet-50 + Axial &  38.8 & - & - & 44.0 & 51.4 \\
\hline
Video K-Net & ResNet50 & 40.5 &  63.5 & 44.5 & 40.7 & 49.9 \\
Video K-Net & Swin-base & 51.4 & 77.2 & 56.1 & 49.0 & 58.4 \\
Video K-Net & Swin-base* & 54.1 & 79.0 & 59.5 & 49.7 & 59.9 \\
\hline
\end{tabular}
}
\end{table}

\begin{table}[!t]
	\centering
	\caption{\small \textbf{Results on VIPSeg-VPS~\cite{miao2022large} validation dataset.} We report VPQ and STQ for reference. Following work~\cite{miao2022large}, we report VPQ score at different window sizes (1,2,4,6).}
	\label{tab:sota_vipseg}
  \scalebox{0.65}{
\begin{tabular}{ r|c|cccccc}
\toprule[0.15em]
 Method& backbone & $VPQ^{1}$ & $VPQ^{2}$ & $VPQ^{4}$ & $VPQ^{6}$ & VPQ & STQ \\
\toprule[0.15em]
VIP-DeepLab~\cite{ViPDeepLab} & ResNet50 & 18.4 & 16.9 & 14.8 & 13.7 & 16.0 & 22.0 \\
VPSNet~\cite{kim2020vps} & ResNet50 & 19.9 & 18.1 & 15.8 & 14.5 & 17.0 & 20.8 \\
SiamTrack~\cite{woo2021learning_associate_vps} & ResNet50 & 20.0 & 18.3 & 16.0 & 14.7 & 17.2 & 21.1 \\
Clip-PanoFCN~\cite{miao2022large} & ResNet50 & 24.3 & 23.5 & 22.4 & 21.6 & 22.9 & 31.5 \\
\hline
Video K-Net & ResNet50 & 29.5 & 26.5 & 24.5 & 23.7 & 26.1 & 33.1 \\
Video K-Net & Swin-base & 43.3 & 40.5 & 38.3 & 37.2 & 39.8 & 46.3 \\
\hline
\end{tabular}
}
\end{table}

\begin{table*}[h!]
    \footnotesize
	\centering
	\caption{\small Ablation studies and comparison analysis on KITTI-STEP validation set. All the experiments use ResNet-50 as backbone.}
    \subfloat[Ablation Study on Each Components.]{
    \label{tab:ablation_a}
	    \begin{tabularx}{0.38\textwidth}{c c c c c c c} 
		        				\toprule[0.15em]
    	baseline & KAE & KL & KF & STQ & AQ & SQ  \\
        \toprule[0.15em]
        K-Net    &  &  &  & 67.5 & 65.5 & 68.9 \\
        & \checkmark &  &  & 69.3 & 69.0 & 69.8 \\
        & \checkmark & \checkmark &  & 70.2 & 71.2 & 69.7 \\
    	& \checkmark & \checkmark & \checkmark & 70.9 & 70.8 & 71.2 \\
        \bottomrule[0.1em]
	    \end{tabularx}
    } \hfill
    \subfloat[Needs of Appearance Embeddings]{
    \label{tab:ablation_b}
		\begin{tabularx}{0.28\textwidth}{c c c} 
			\toprule[0.15em]
			Method & AQ & STQ \\
			\midrule[0.15em]
            RoI-Align~\cite{qdtrack} & 68.8 & 69.1 \\
            Mask-Emb~\cite{woo2021learning_associate_vps} & 67.3 & 68.1 \\
            Ours  & 70.8 & 70.9 \\
             Ours + Mask-Emb~\cite{woo2021learning_associate_vps}  & 70.3 & 70.8 \\
			\bottomrule[0.1em]
		\end{tabularx}
    } \hfill
    \subfloat[Effect of sampling in association.]{
    \label{tab:ablation_c}
		\begin{tabularx}{0.30\textwidth}{c c c c} 
			\toprule[0.15em]
			Method & STQ & AQ & SQ  \\
			\midrule[0.15em]
			 K-Net  & 67.5 & 65.5 & 68.9 \\
            GT-based (ours) & 69.3 & 69.0 & 69.8 \\
            sampling in~\cite{qdtrack} & 63.1 & 62.1 & 64.3 \\
			\bottomrule[0.1em]
		\end{tabularx}
    } \hfill
    \vspace{2mm}
    \subfloat[Ablation Study on Linking and Fusing Stage.]{
     \label{tab:ablation_d}
	    \begin{tabularx}{0.25\textwidth}{c c c c} 
		        				\toprule[0.15em]
    		 Stage & STQ & AQ & SQ  \\
    		\toprule[0.15em]
    	    3 & 70.9 & 70.8 & 71.2 \\
    	    2 & 68.5 & 68.2 & 69.3 \\
    	    1 & 66.9 & 63.4 & 67.3 \\
        	\bottomrule[0.1em]
	    \end{tabularx}
    } \hfill
    \subfloat[Ablation Study on Training Settings ]{
     \label{tab:ablation_e}
	    \begin{tabularx}{0.35\textwidth}{c  c  c c} 
		        				\toprule[0.15em]
    		 Settings  & STQ & AQ  & SQ \\
    		\toprule[0.15em]
    	    joint training  &  70.9 & 70.8 & 71.2\\
    	   only train the key frame & 70.1 & 70.1 & 69.8 \\
        	\bottomrule[0.1em]
	    \end{tabularx}
    } \hfill
    \subfloat[Ablation Study on Kernel Fusing]{
     \label{tab:ablation_f}
	    \begin{tabularx}{0.28\textwidth}{c c c c} 
		        				\toprule[0.15em]
    		 Settings  &  STQ & AQ & SQ \\
    		 \toprule[0.15em]
    		   K-Net & 67.5 & 65.5 & 68.9 \\
    	       w Update & 70.9 & 70.8 & 71.2 \\
    	       w/o Update & 67.1 & 66.2 & 68.3 \\
        	\bottomrule[0.1em]
	    \end{tabularx}
    } \hfill
\end{table*}

\subsection{Main Results}

\noindent
\textbf{Results on KITTI-STEP.} As shown in Tab.~\ref{tab:kitti_baselines}, our method achieves 0.71 STQ and 0.46 VPQ using ResNet50 as backbone and achieves the new state-of-the-art results among previous works. After applying a stronger backbone~\cite{liu2021swin}, we obtain about 3\% gains on STQ and 7\% gains on VPQ. This suggests strong feature representation leads to better segmentation-level prediction than pixel-level prediction. Compared with VPSNet~\cite{kim2020vps} and Motion-Deeplab~\cite{STEP}, our method does not need post-processing steps or extra optical flow to warp features. Compared with Motion-Deeplab~\cite{STEP}, our method outperforms it by 13\% and 7\% on validation set and test set respectively. The results suggest the suitability of our framework to serve as a new and stronger baseline. We find 0.5\% noise on STEP validation set.

\noindent
\textbf{Results on Cityscape-VPS.} Tab.~\ref{tab:city_vps_results} compares our method with previous works on Cityscapes-VPS datasets. For ResNet50, our method outperforms VPSNet~\cite{kim2020vps} by 0.3\%. Moreover, we follow the same pretraining procedure of ViP-Deeplab~\cite{ViPDeepLab} and achieve 62.2\% VPQ. Our method with Swin-Transformer base achieves better results than ViP-Deeplab, which is trained with stronger data augmentation~\cite{cubuk2018autoaugment}. Note that we find 0.5\% noise for this dataset. Moreover, we do not use the kernel fusion on this dataset due to the low frame rate of Cityscapes-VPS.

\noindent
\textbf{Results on VSPW.} We further carry out experiments on VSPW dataset for VSS task to prove the generalization of Video K-Net. All the methods are re-implemented on our codebase and achieve higher results than the original paper~\cite{miao2021vspw}. As shown in Tab.~\ref{tab:vspw}, Video K-Net boosts Deeplabv3+~\cite{deeplabv3plus} and PSPNet~\cite{zhao2017pyramid} by a significant margin on both mIoU (2-3\%) and $mVC$ (3-4\%). This proves the generalization ability of both kernel fusion and kernel linking.

\noindent
\textbf{Results on Youtube-VIS-2019.} In Tab.~\ref{tab:sota2019_vis}, we report the results on Youtube-VIS datasets. Compared with previous work VisTR~\cite{VIS_TR}, our method achieves better results (4\%) but with less training times (12 epochs vs 300 epochs).

\noindent
\textbf{Results on VIPSeg-VPS.} In Tab.~\ref{tab:sota_vipseg}, we further report the results on recently challenging VPS dataset VIPSeg~\cite{miao2022large}. Using ResNet50 backbone, our Video K-Net outperforms Clip-PanoFCN~\cite{miao2022large} by 3\% VPQ and 1.6\% STQ with only 12 epoch training (half training time of Clip-PanoFCN). We further use a larger model (Swin-base), which results in new state-of-the-art results and outperforms previous work by 16.6\% VPQ and 14.8\% STQ.

\subsection{Ablation Study}
\label{sec:ablation}

\noindent
\textbf{Ablation Study on Each Component.} In Tab.~\ref{tab:ablation_a}, we first perform ablation studies on the effectiveness of each component. Adding KAE yields 2.8\% STQ improvements with more significant gains on AQ (about 2.4\%). It shows that directly learning kernel association is a key factor for tracking. Adding Kernel Linking module (KL) further improves AQ by 1.0\%. Finally, appending kernel fusing results in 0.7\% STQ improvement and 1.4 \% improvement on SQ. The AQ decreases a little, mainly because both KL and KF are fighting for tracking quality and segmentation quality, respectively.

\noindent
\textbf{Needs of Extra Appearance Embedding.} In Tab.~\ref{tab:ablation_b}, we present detailed comparison with recent box based appearance embedding and mask based appearance embedding. In particular, we re-implement their methods on our baseline K-Net. We also use the same embedding loss functions: Equ.~\ref{equ:qd_loss} and Equ.~\ref{equ:qd_loss_aux} for fair comparison. From the table, our simple design leads to the best result. From the last row of Tab.~\ref{tab:ablation_b}, even combing both the appearance embedding and kernels, there is no clear gain. We implement this by adding mask-grouped appearance features from backbone and kernel embeddings together. That indicates kernels have already encoded the discriminative information, which is consistent with our toy experiments results: the pure kernel information is good enough for tracking. 

\noindent
\textbf{Effect of Sampling Examples in Association.} Pang~\etal~\cite{qdtrack} employ RPN to generate large samples to improve the tracking results. However, we find that directly using this method into our framework leads to inferior results because most kernels are not used to generate masks with Hungarian assignment strategy during the training stage. Adding more samples brings noise to the segmentation prediction shown in Tab.~\ref{tab:ablation_c}, causing a marked decrease in SQ metric. Thus, we adapt to assign training kernel samples that are matched with ground truth masks. The matched kernels are more robust during training.

\noindent
\textbf{Ablation on Link Stage.} In Tab.~\ref{tab:ablation_d}, we present ablation on the stage of linking kernels, where we find linking kernels at the last stage achieves the best results. Early fusing of kernels also leads to bad results, since initial kernels are not very accurate. This results in bad misalignment on both kernel and kernel features.

\begin{figure}[!t]
	\centering
	\includegraphics[width=0.80\linewidth]{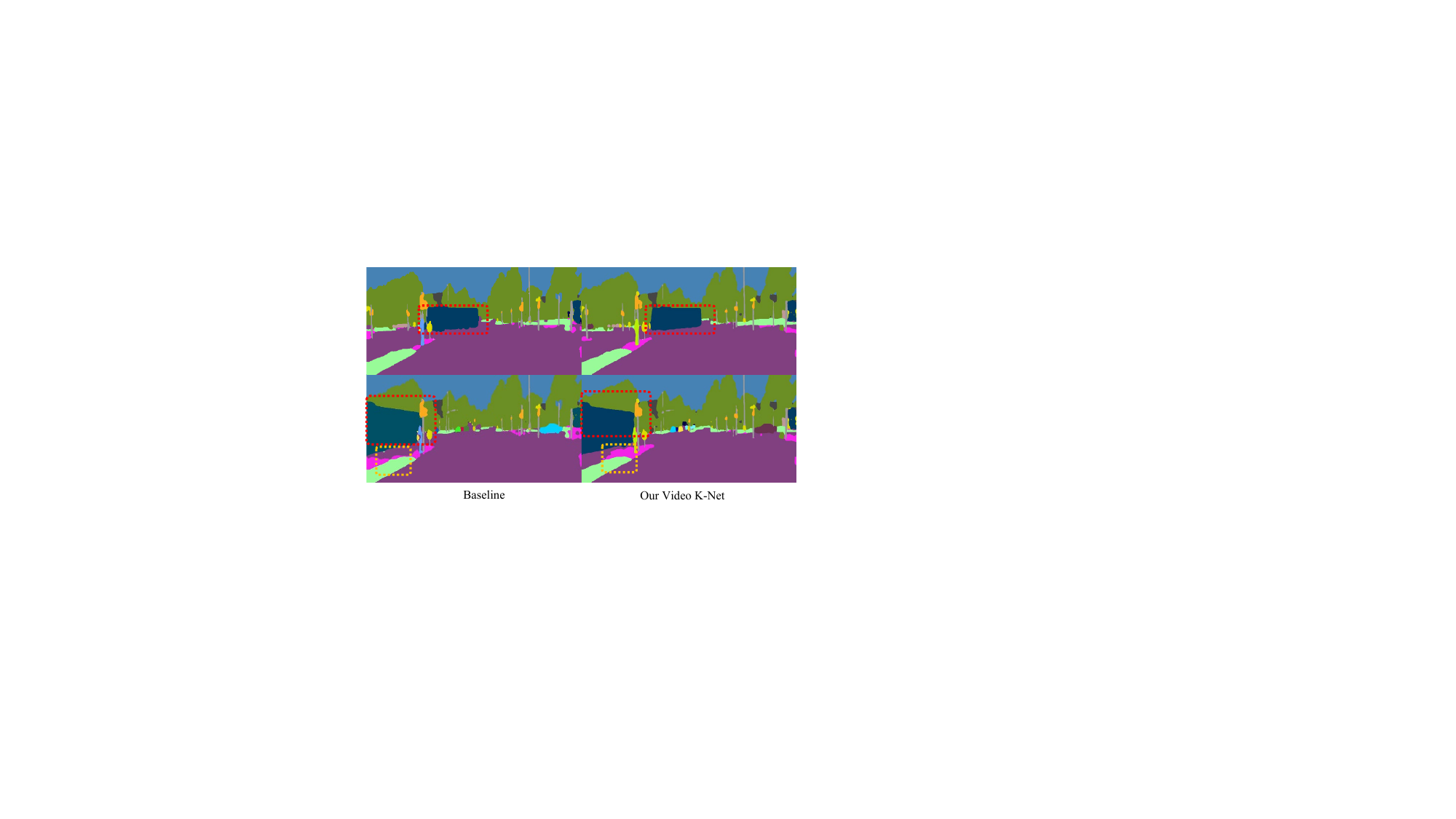}
	\caption{\small Visualization of improvements over baseline methods. Tracking improvement in red boxes and segmentation improvement in yellow boxes. Best viewed in color and by zooming in. }
	\label{fig:vis_baseline}
\end{figure}

\vspace{1mm}
\noindent
\textbf{Effect on Training Settings.} Moreover, in Tab.~\ref{tab:ablation_e}, we find joint training with both frames $I_{key}$ and $I_{ref}$ performs better than only training $I_{key}$. This is mainly because both the proposed Kernel Linking and Kernel Fusion need to learn to group similar kernels and separate the different kernels.

\begin{figure}[!t]
	\centering
	\includegraphics[width=1.0\linewidth]{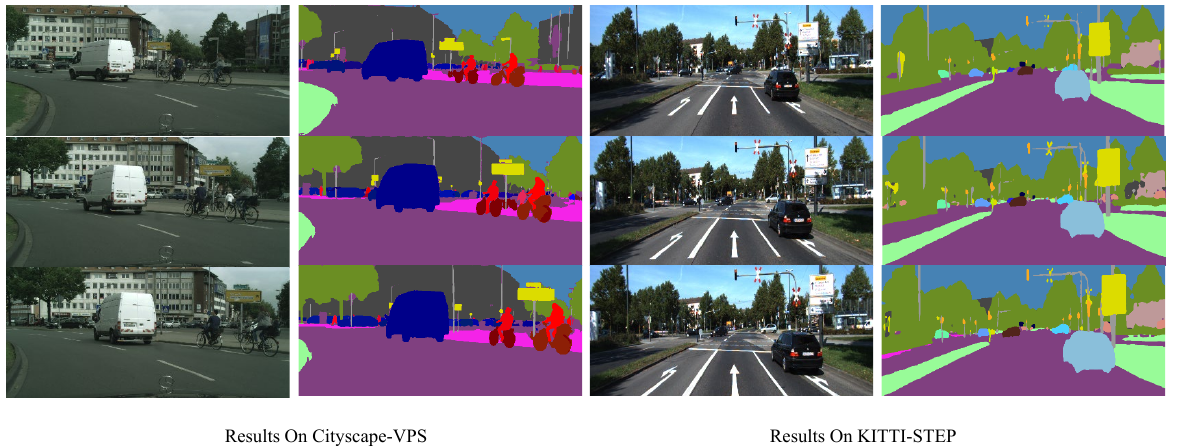}
	\caption{\small Visual Examples of our Video K-Net. The left is on Cityscapes-VPS validation set, while the right is on KITTI-STEP validation set (video sequences from top to down). The same instances are shown with the same color. Best view it on screen. }
	\label{fig:results_vis}
\end{figure}

\noindent
\textbf{Ablation on Kernel Fusing Design.} In Tab.~\ref{tab:ablation_f}, we find the importance of Kernel Updating module in Kernel Fusing. Since previous features are not aligned to the current frame, we use current-frame features to re-weight previous kernels to filter out noise and outliers in the previous kernel.

\subsection{Visualization and More Analysis}

\noindent
\textbf{Visualization over Baseline.} In Fig.~\ref{fig:vis_baseline}, we visualize the improvements over the K-Net baseline. The red boxes show the tracking consistency, while the yellow boxes show the segmentation consistency. Our Video K-Net achieves better results on both tracking and segmentation qualities.

\noindent
\textbf{More Qualitative Results.} In Fig.~\ref{fig:results_vis}, we present several visual results on Citscapes-VPS and KITTI-STEP datasets. We find most foreground objects (person and car) can be well tracked and segmented. \textit{More visual examples can be found in the appendix file}.

\noindent
\textbf{Parameter and GFLOPs} Compared with K-Net baseline, our method only adds about \textbf{2.3\% GFLOPs} and \textbf{1.8\% Parameters} given $800\times1333$ image inputs. 

\noindent
\textbf{Limitation and Future Work.} One limitation of Video K-Net is that we only explore one frame during both training and inference. This setting is mainly for fair comparison with other works~\cite{STEP,kim2020vps}. Both exploring the \textit{multiple frames} information via kernels and handling on long video inputs will be our future work.
\section{Conclusion}
\label{sec:conclusion} 
We present \textbf{Video K-Net} a simple, strong and unified system for fully end-to-end video panoptic segmentation. We propose to learn the kernel association embeddings directly from the encoded `thing' kernels. Then we link and fuse the kernels to jointly improve both tracking and segmentation results. Despite simplicity, our method achieves state-of-the-art results on three VPS datasets including Cityscapes-VPS, KITTI-STEP and recently new proposed VIPSeg datasets. In particular, our method boosts previous methods over 10\% improvements on KITTI-STEP and Cityscapes-VPS datasets. In particular, compared to previous works, our method has a much simpler pipeline but achieves better results along with less computation cost. We also show the generalization ability on the VSPW dataset for VSS task and competitive results on YouTub-VIS dataset for VIS task. We hope our proposed Video K-Net would be a simple and strong baseline and benefit unified video segmentation field.

\noindent
\textbf{Acknowledgement.} This study is partly supported under the RIE2020 Industry Alignment Fund Industry Collaboration Projects (IAF-ICP) Funding Initiative, as well as cash and in-kind contribution from the industry partner(s). It is also partially supported by the NTU NAP grant. This research is also supported by the National Key Research and Development Program of China under Grant No. 2020YFB2103402. We also thank for the GPU resource provided by SenseTime Research.

{\small
\bibliographystyle{ieee_fullname}
\bibliography{egbib}
}
In this report, we provide the following information in addition to the
main paper: more experimental details and more visualization results on Cityscapes-VPS~\cite{kim2020vps} and KITTI-STEP~\cite{STEP}. All the model use AdamW for training.

\section{More Experimental Details}

\noindent
\textbf{Detailed pretraining setting.} 

For COCO~\cite{coco_dataset} dataset pretraining, all the models are trained following original K-Net settings~\cite{zhang2021knet}. We adopt the multi-scale training setting as previous work~\cite{detr} by resizing the input images such that the shortest side is at least 480 and at most 800 pixels, while the longest size is at most 1333. We also apply random crop augmentations during training, where the train images are cropped with probability 0.5 to a random rectangular patch which is then resized again to 800-1333. All the models are trained for 36 epochs.

For Mapillary~\cite{neuhold2017mapillary} dataset pretraining, we mainly follow the Panoptic-Deeplab setting~\cite{cheng2020panoptic}. We adopt the multi-scale training where the the scale ranges from 1.0 to 2.0 of origin image size, then we apply a random crop of 1024 $\times$ 2048 patches. The horizontal flip is applied. The pretraining process takes 240 epochs. The Mapillary pretraining is for fair comparison, since the ViP-Deeplab~\cite{ViPDeepLab} all use the Mapillary pretraining for better results. Note that, we \textit{only} pretrain our largest models with Swin-base~\cite{liu2021swin} as backbone for fair comparison with previous work.

\noindent
\textbf{Training and inference on Cityscapes-VPS.}

For Cityscapes-VPS training, we follow previous VPSNet~\cite{kim2020vps} that we randomly sample one frame from the nearest one frame as the reference frame. We adopt the multi-scale training where the scale ranges from 1.0 to 2.0 of origin images size then we apply a random crop of 1024 $\times$ 2048 patches. For the Swin-base model, we apply a random crop of 800 $\times$ 1600 patches to save memory and computation cost. The total training epoch is set to 8. 

During the inference, the previous frame plays as the reference frame, the kernel information is directly propagated into the next frame in an online manner.

\noindent
\textbf{Training and inference on KITTI-STEP.}

For KITTI-STEP training, we follow previous Motion-Deeplab~\cite{STEP} that we randomly sample one frame from near \textit{three} frames as the reference frame. We adopt the multi-scale training where the scale ranges from 1.0 to 2.0 of origin images size, then we apply a random crop of 384 $\times$ 1248 patches. The total training epoch is set to 12. The inference procedure is the same as Cityscapes-VPS dataset. Following~\cite{STEP}, we also use Cityscapes pretraining before training on STEP, which leads to about 3\% STQ gain on the K-Net baseline.

\noindent
\textbf{Training and inference on VSPW.}

For VSPW dataset, we adopt the same setting on training K-Net~\cite{zhang2021knet} on ADE-datasets~\cite{ADE20K}. We randomly sample one frame from the nearest three frames as the reference frame. The inference procedure is the same as Cityscapes-VPS dataset.

\noindent
\textbf{Training and inference on VIPSeg.}

For VIPSeg dataset, we adopt the same setting on of KITTI-STEP. In particular, we use COCO-pretrained K-Net following~\cite{miao2022large}. The entire training time is 12 epochs. We adopt the multi-scale training where the scale ranges from 1.0 to 2.0 of origin images size, then we apply a random crop of 720 $\times$ 720 patches. The inference procedure is the same as Cityscapes-VPS dataset.

\noindent
\textbf{Training and inference on YoutubeVIS-2019.} 
For YoutubeVIS-2019 dataset, we adopt the same training pipeline as VisTR~\cite{VIS_TR} by sampling five frames to train Video K-Net jointly. We adopt COCO-pretrained model for initialization and train the video for only 12 epochs.

\section{More Visualization Results}

\begin{figure*}[!t]
	\centering
	\includegraphics[width=0.90\linewidth]{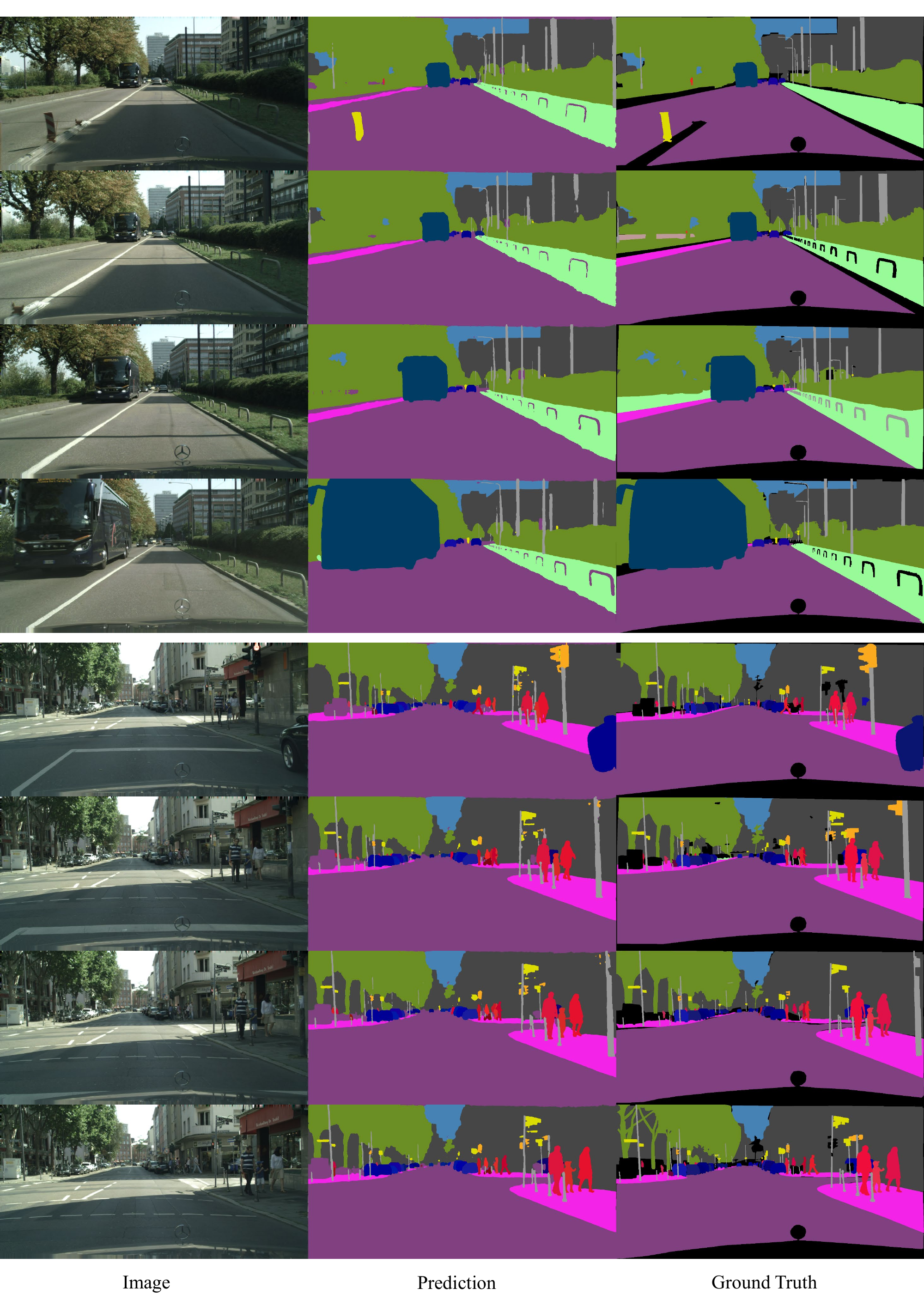}
	\caption{More visual examples of our Video K-Net on CityScapes-VPS dataset. The same instances are shown with the same color. Best view it on screen. }
	\label{fig:more_vis_city}
	
\end{figure*}

\begin{figure*}[!t]
	\centering
	\includegraphics[width=0.95\linewidth]{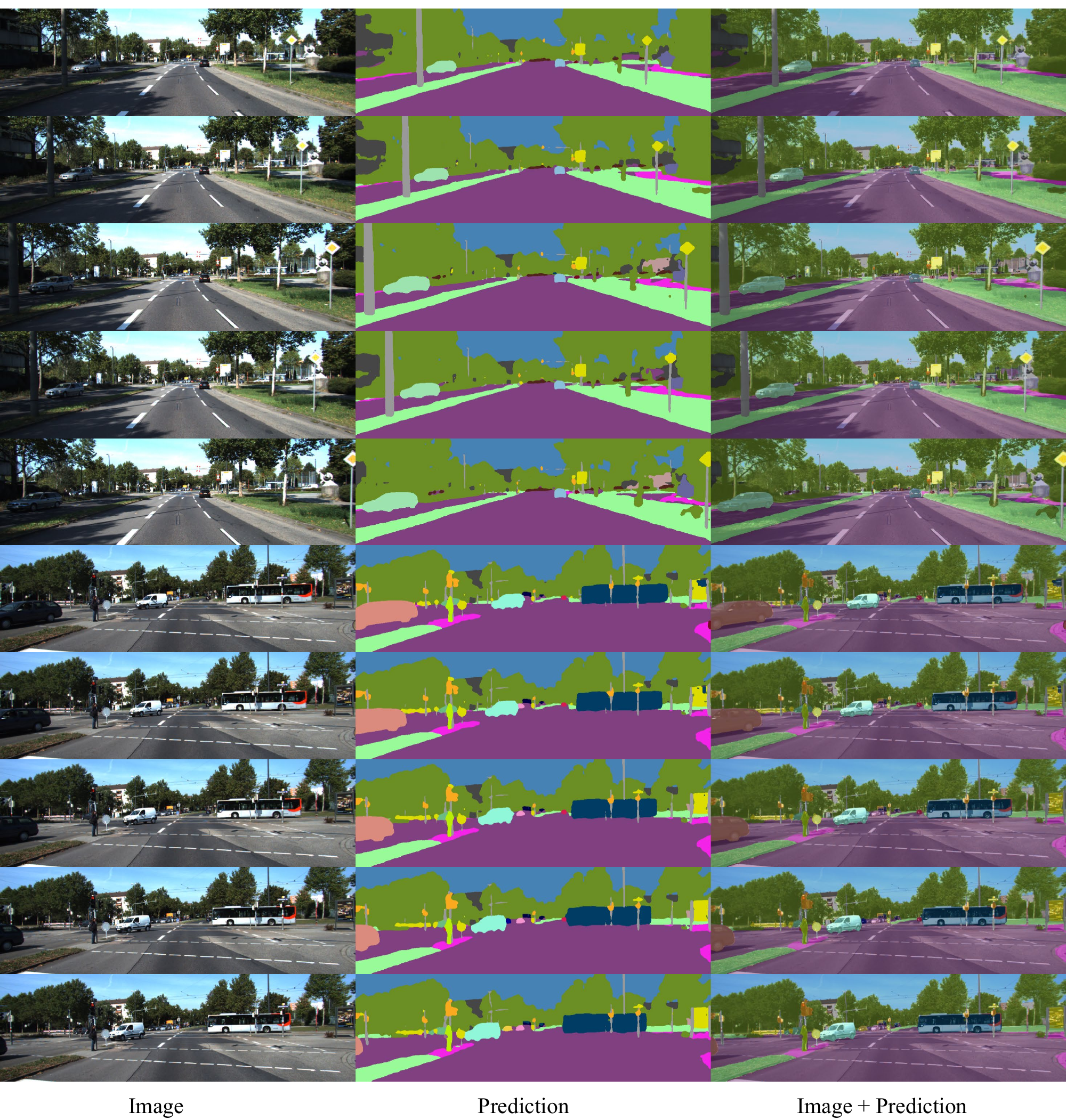}
	\caption{More visual examples of our Video K-Net on STEP dataset. The same instances are shown with the same color. Best view it on screen. }
	\label{fig:more_vis_step}
	
\end{figure*}

\begin{figure*}[!t]
	\centering
	\includegraphics[width=0.95\linewidth]{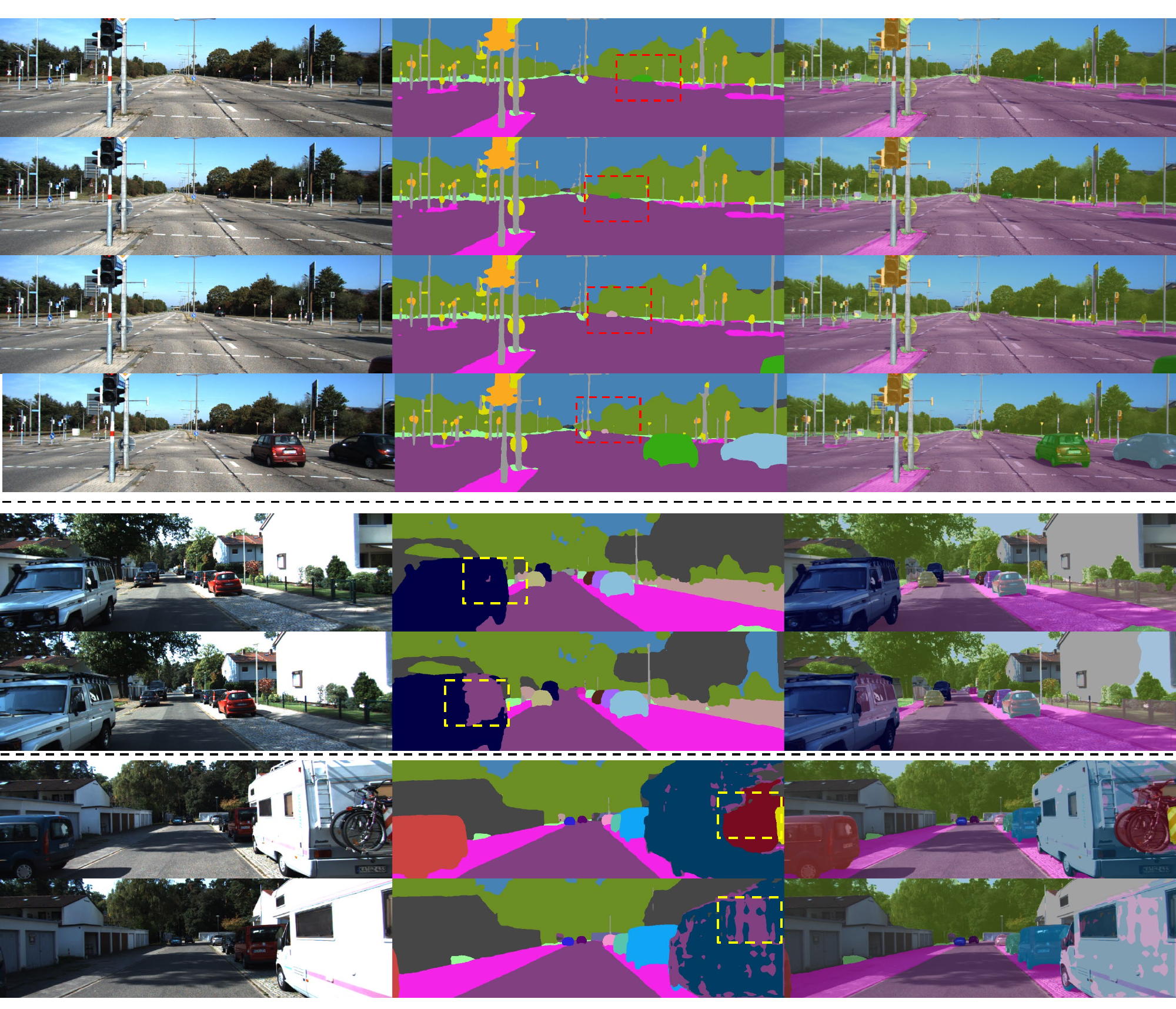}
	\caption{
	Failure cases of Video K-Net on STEP dataset. The same instances are shown with the same color. The red boxes show the tracking errors, while the yellow boxes show the segmentation errors. Best view it on screen. }
	\label{fig:vis_step_failure}
	
\end{figure*}
\noindent
\textbf{More Visualization on Cityscapes-VPS.}
In Fig.~\ref{fig:more_vis_city}, we present more visual examples on Cityscaeps VPS dataset. Compared with Ground Truth, our method can segment and track well for each pixel for various scale object inputs. We use the Swin-base model for visualization.

\noindent
\textbf{More Visualization on KITTI-STEP.}
In Fig.~\ref{fig:more_vis_step}, we give more visual results on the KITTI-STEP validation set. On both clips, Video-KNet shows convincing results.

\noindent
\textbf{Failure Cases Analysis.}
In Fig.~\ref{fig:vis_step_failure}, we present three failure cases of our Video K-Net. The first case is the tracking failure case where the car moves fast in remote scene and then there is an ID switch. We believe adding motion cues will improve this case. The last two cases show the segmentation errors. The first is because the color of cars' window is similar to the ground. The second is caused by less training cases: bike on the truck, which makes network hard to predict.  

\noindent
\textbf{Border Impact.} Our work pushes the boundary of video segmentation algorithms through simplicity and effectiveness. Since most applications are the video input, this work could also ease and accelerate the model production in real-world applications, such as in autonomous driving. Due to the limited dataset size, we do not evaluate the robustness of the proposed method on corrupted video inputs or extremely long video inputs. 

\end{document}